\documentclass[letterpaper, 10 pt, conference]{ieeeconf}  

\IEEEoverridecommandlockouts                              

\usepackage{graphicx}
\usepackage{epsfig} 
\usepackage{mathptmx} 
\usepackage{times} 
\usepackage{amsmath} 
\usepackage{amssymb}  
\usepackage{adjustbox}
\usepackage{subcaption}
\usepackage{multirow}
\usepackage{hyperref}
\usepackage{xcolor}
\usepackage{cite}
\usepackage{colortbl}
\usepackage{dirtytalk}
\usepackage[normalem]{ulem}

\title{\LARGE \bf
Autonomous Dissection in Robotic Cholecystectomy
}

\author{Ki-Hwan Oh$^{1}$,  Leonardo Borgioli$^{1}$,  Milo\v s \v Zefran$^{1}$, Valentina Valle$^{2}$ and Pier Cristoforo Giulianotti$^{2}$
\thanks{*This work was not supported by any organization}
\thanks{$^{1}$K.H. Oh, L. Borgioli and Milo\v s \v Zefran are with the Robotics Lab,  Department of Electrical and Computer Engineering, College of Engineering, University of Illinois Chicago, Chicago, IL 60607, USA.}%
\thanks{$^{2}$V. Valle and P.C. Giulianotti are with the Surgical Innovation and Training Lab,  Department of Surgery, College of Medicine, University of Illinois Chicago, Chicago, IL 60607, USA.}%
}

\begin{document}

\maketitle
\thispagestyle{empty}
\pagestyle{empty}

\begin{abstract}

Robotic surgery offers enhanced precision and adaptability, paving the way for automation in surgical interventions. Cholecystectomy, the gallbladder removal, is particularly well-suited for automation due to its standardized procedural steps and distinct anatomical boundaries. A key challenge in automating this procedure is dissecting with accuracy and adaptability. This paper presents a vision-based autonomous robotic dissection architecture that integrates real-time segmentation, keypoint detection, grasping and stretching the gallbladder with the left arm, and dissecting with the other arm. We introduce an improved segmentation dataset based on videos of robotic cholecystectomy performed by various surgeons, incorporating a new ``liver bed'' class to enhance boundary tracking after multiple rounds of dissection. Our system employs state-of-the-art segmentation models and an adaptive boundary extraction method that maintains accuracy despite tissue deformations and visual variations. Moreover, we implemented an automated grasping and pulling strategy to optimize tissue tension before dissection upon our previous work. Ex vivo evaluations on porcine livers demonstrate that our framework significantly improves dissection precision and consistency, marking a step toward fully autonomous robotic cholecystectomy.

\end{abstract}


\section{Introduction}

Robotic-assisted surgery (RAS) continues to revolutionize modern medicine by enabling higher precision, reduced patient trauma, and faster recovery than traditional open or laparoscopic procedures. This favorable setting has spurred research into automating various surgical tasks such as tissue ablation \cite{ayvali2016using}, blunt dissection~\cite{blunt_dissection, nagy2018ontology}, and suturing~\cite{Jackson2016, Lu2022autosuture, iyer2013single, sen2016suture, Tracking_thread,ostrander2024current}. For instance, the work in \cite{Sagitov2018} demonstrated an algorithm for automated suturing based on visual tracking. At the same time, point cloud-based methods have been explored to model soft tissue deformations~\cite{blunt_dissection}, estimate applied forces~\cite{gao2018learning,pan5115117review}, and for surgical planning~\cite{lu2021spatial}.

Most prior approaches to task automation in RAS rely on endoscopic imaging, which poses challenges due to its two-dimensional representation and variable lighting conditions. Advanced segmentation and detection techniques are critical for overcoming these limitations. Recent breakthroughs bring new frameworks for subtask automations \cite{framework_subtask}, with, for example, endoscopic servo control \cite{visualservoingcamera}. However, their methods often remain confined to pixel-level processing without establishing logical tissue connections~\cite{chiranjeevi2024intelligent}, or relying on markers \cite{d2021autonomous}.

Among various interventions, cholecystectomy, the surgical removal of the gallbladder attached to the liver, has emerged as an ideal testbed for automation due to its relatively straightforward anatomy and standardized sequence of surgical steps. Previously~\cite{oh2024framework}, we introduced a framework to automate the control of a single da Vinci surgical robot instrument relying only on endoscopic images, like surgeons, to dissect\footnote{In the surgical literature, dissection refers to the actual separation of tissues; in our work, the term refers to energy delivery to a particular location on a specimen.} along the trajectory between the porcine gallbladder and the liver. However, we faced some limitations: (a) the dataset we generated was preliminary, and thus, the trained model could not perform as expected under varying conditions, such as changes in tissue colors over time and variations in the size and shape of the gallbladder; and (b) the dissection trajectory was planned offline, but we discovered that the tissue deformed after the energy was applied even though the tissues were not touched with other instruments.

To address these shortcomings, we present a new segmentation dataset based on the Comprehensive Robotic Cholecystectomy Dataset (CRCD)~\cite{oh2024crcd}, which includes robotic cholecystectomy procedures performed with the classic da Vinci system. In addition, we introduce a new class for tissue segmentation that corresponds to the region where the gallbladder used to be attached to the liver (\textit{liver bed}), so the model can track the boundary even after multiple rounds of dissection. Furthermore, we train several state-of-the-art models for tissue instance segmentation, such as the MaskDINO~\cite{li2022mask} and YOLO11~\cite{yolo11_ultralytics}, and compare the performance with the previous Detectron2~\cite{wu2019detectron2} model. We also develop an online boundary extraction method and an improved method for following the boundary. Finally, we propose a novel method to automatically grasp and pull the tissue before dissection. Together, the ability to track the boundary online and grasp the gallbladder represents a major step towards making our automated procedure much closer to how surgeons perform dissection. The effectiveness of our framework is successfully evaluated ex vivo using porcine liver.

Our contributions are twofold: (a) developing a comprehensive vision-based control framework that integrates online segmentation and keypoint detection for precise dissection along tissues that deform in time; (b) a framework that allows the robot to automatically grasp the gallbladder to maximize tissue tension and dissect the gallbladder from the liver, adapting to tissue deformations; and (c) an ex vivo evaluation that demonstrates the performance of the system in a realistic surgical scenario.




\section{Expanded Dataset}

\subsection{Revised Custom Segmentation Dataset}

One limitation of our previous dataset was the lack of variability in gallbladder and liver features, such as differences in shape, size, and color. Additionally, we observed that the model struggled to track deformed tissues after monopolar energy was applied via the electrosurgical unit, as the dissected region no longer resembled the original gallbladder or liver (Fig.~\ref{fig:bnd_after_dissect}). These limitations prevented the robot from performing multiple rounds of dissection, which is necessary for the complete separation of the gallbladder from the liver.

\begin{figure}[ht]
    \centering
    \includegraphics[width=\linewidth]{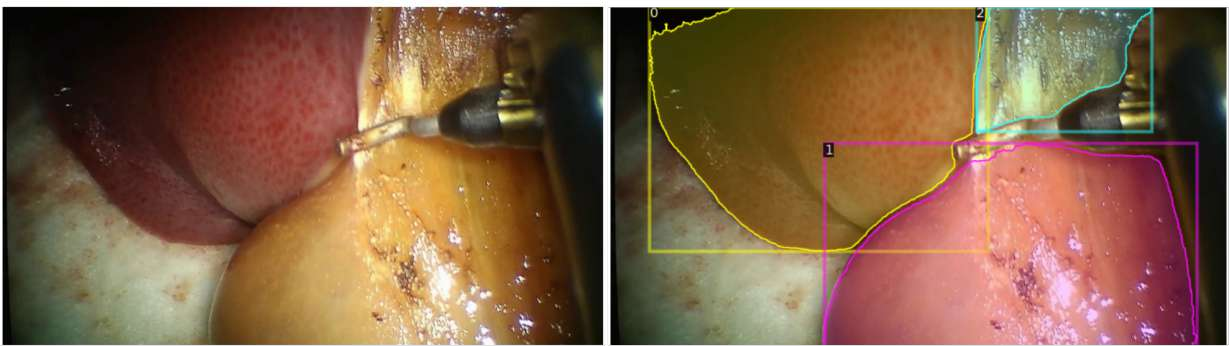}
    \caption{Example of tissue segmentation annotation to train the model. The tissue is Liver (Orange), Gallbladder (Pink), and Liver Bed (Green) }
    \label{fig:seg_example}
\vspace{-2mm}
\end{figure}

To address these challenges, we enriched the segmentation dataset by introducing a new label, liver bed, which represents the liver surface where the gallbladder was previously attached (Fig.~\ref{fig:seg_example}). We used the Segment Anything Model 2 (SAM2)~\cite{ravi2024sam} to annotate videos from the Comprehensive Robotic Cholecystectomy Dataset (CRCD)~\cite{oh2024crcd}, a large-scale ex vivo robotic cholecystectomy dataset featuring procedures performed by multiple surgeons. This dataset closely aligns with our intended application and captures the dynamic tissue variations during surgery. SAM2 segments objects in single images or sequential video frames using prompts such as pixel-based cues or bounding boxes. However, SAM2 does not inherently classify the segmented regions, so we manually labeled each tissue category to create a structured training dataset. This allowed us to fine-tune state-of-the-art instance segmentation models, including YOLO11, for improved tissue recognition.

The number of annotations of the new segmentation dataset is shown in Table~\ref{tab:customdata_seg}. These improvements contribute to a more robust dataset that better represents real-world surgical conditions, facilitating the development of more accurate and reliable segmentation models.


\begin{table}[ht]
\resizebox{\columnwidth}{!}{%
\begin{tabular}{l|l|cc|cc}
\hline \hline
\multirow{2}{*}{\textbf{Data Type}}    & \multirow{2}{*}{\textbf{Categories}} & \multicolumn{2}{l|}{\textbf{Previous~\cite{oh2024framework}}}     & \multicolumn{2}{c}{\textbf{New}}                             \\ \cline{3-6} 
                              &                             & \multicolumn{1}{l|}{Train} & Test & \multicolumn{1}{c|}{Train} & Test \\ \hline \hline
\multirow{3}{*}{Segmentation} & Pig Liver                   & \multicolumn{1}{c|}{1430}                       & 356  & \multicolumn{1}{c|}{25988}          & 8690          \\
                              & Pig Gallbladder             & \multicolumn{1}{c|}{1429}                       & 359  & \multicolumn{1}{c|}{25988}          & 8690          \\
                              & Liver Bed                   & \multicolumn{1}{c|}{-   }                       & -    & \multicolumn{1}{c|}{21660}          & 7261          \\ \hline
\multirow{2}{*}{Keypoints}    & FBF                         & \multicolumn{1}{c|}{471 }                       & 130  & \multicolumn{1}{c|}{5476}           & 1372          \\
                              & PCH                         & \multicolumn{1}{c|}{1904}                       & 477  & \multicolumn{1}{c|}{7320}           & 1831          \\ \hline \hline
\end{tabular}%
}
\caption{Description of the annotated dataset. Keypoints were annotated for the Fenestrated Bipolar Forceps (FBF) and Permanent Cautery Hook (PCH).}
\label{tab:customdata_seg}
\vspace{-2mm}
\end{table}

\subsection{Keypoints}

\begin{figure}[ht]
    \centering
    \includegraphics[trim={0.0 2.5cm 0.0 2.5cm}, clip, width=0.545\linewidth]{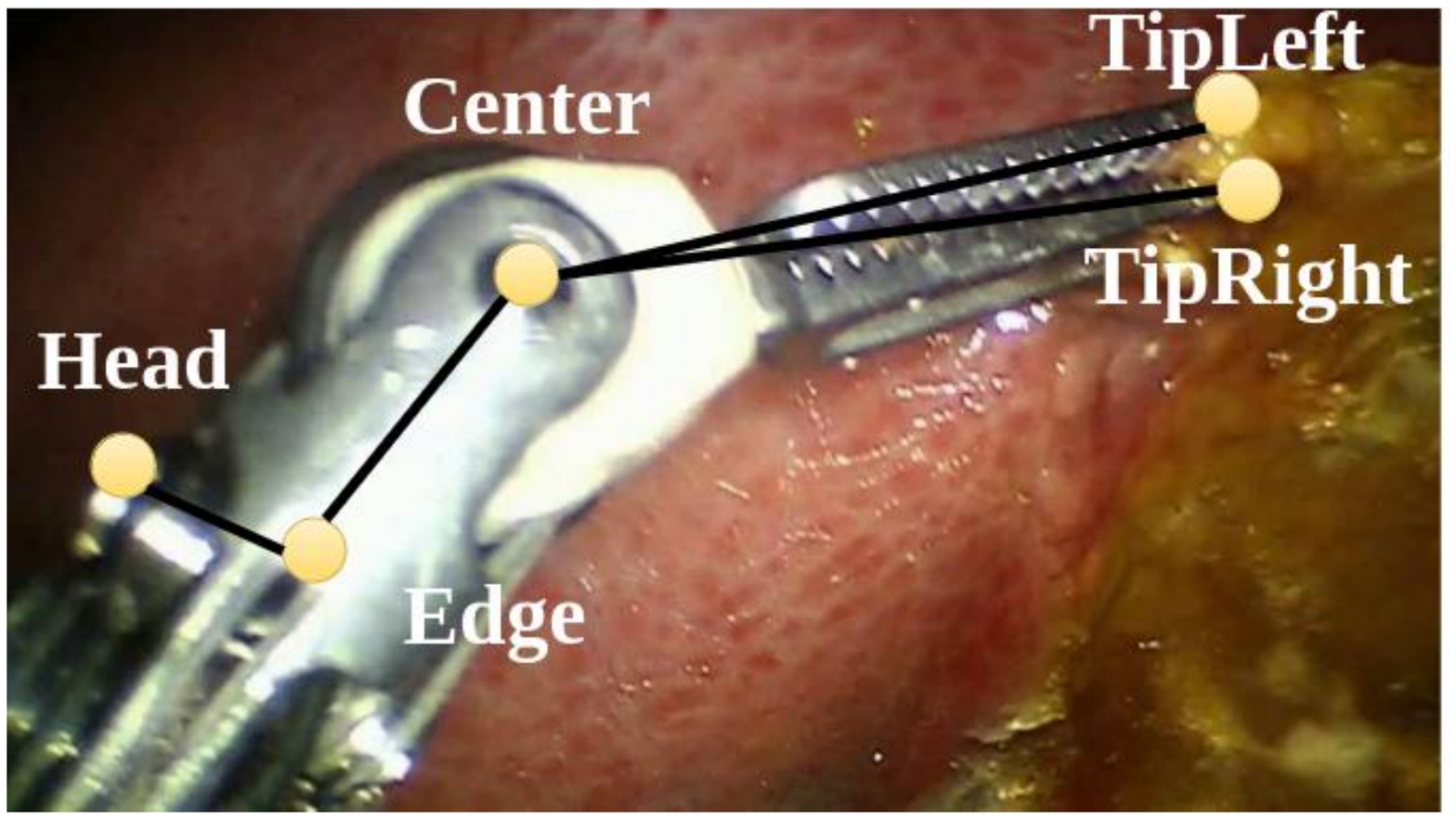}
    \includegraphics[trim={0.0 0.5cm 0.0 0.5cm}, clip, width=0.44\linewidth]{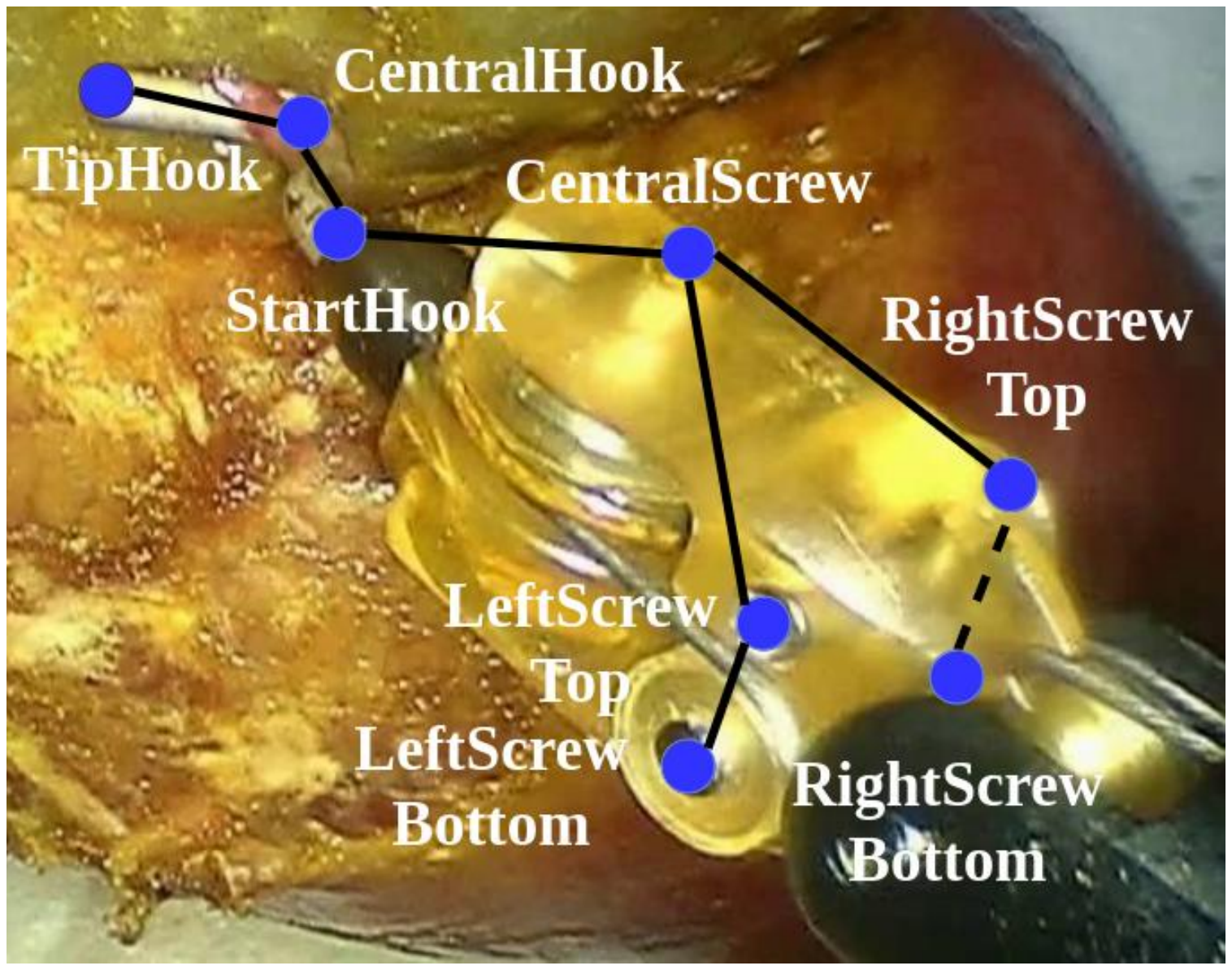}
    \caption{Keypoints structure for each instrument. Fenestrated Bipolar Forceps (FBF, left). Permanent Cautery Hook (PCH, right).}
    \label{fig:kpt_anns}
\vspace{-3mm}
\end{figure}

We also expanded the keypoint dataset to have stronger keypoint tracking of the instruments. The annotations for surgical instruments, including Fenestrated Bipolar Forceps (FBF) and Permanent Cautery Hook (PCH), were manually performed using the COCO annotator~\cite{cocoannotator}. Moreover, we used a different keypoint structure for each instrument to ensure robustness against common transformations. Fig.~\ref{fig:kpt_anns} illustrates the keypoint locations of the FBF and PCH. These keypoints are strategically placed on instrument regions with distinct colors and edges to optimize detection accuracy. Table~\ref{tab:customdata_seg} details the number of annotated instances for the extended instrument keypoint dataset. Both the segmentation and keypoint datasets follow Microsoft's COCO format~\cite{mscoco}, ensuring compatibility with diverse object detection models. The annotated datasets are included in the Expanded CRCD~\cite{oh2024expanded}, which provides additional details and is publicly available.

\begin{figure*}[t]
    \centering
    \includegraphics[scale=0.5]{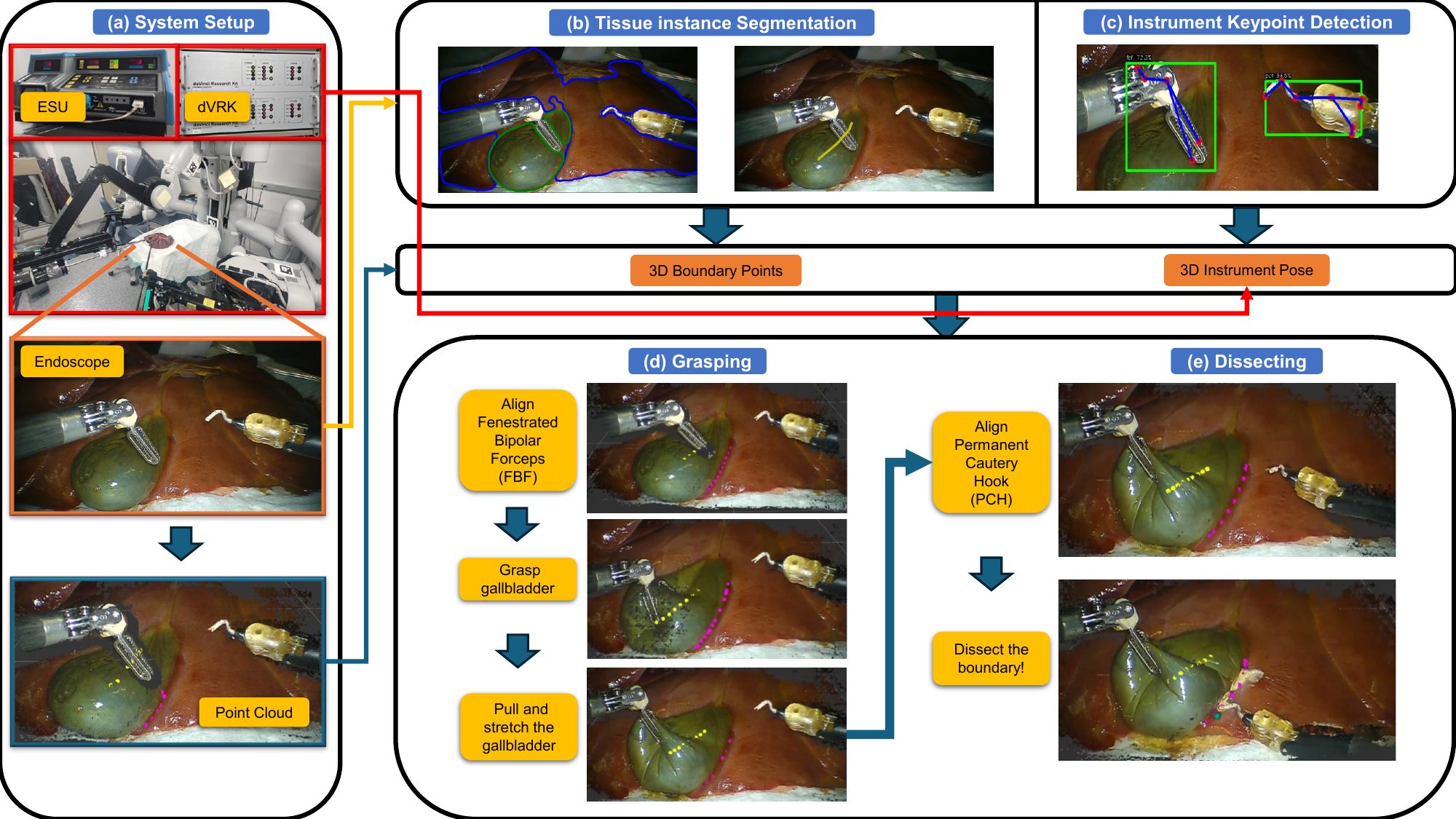}
    \caption{The overall system architecture: (a) Hardware setup: in addition to the dVRK, the ESU is included. Stereo endoscopic images are used to generate real-time 3D point clouds. (b) \& (c) Outputs of the perception models: tissue instance segmentation and instrument keypoint detection. These results are used to compute the 3D dissection boundary and the 3D pose of each surgical instrument. (d) Grasping mechanism using the FBF (Section~\ref{sec:grasp}). (e) Updated dissecting mechanism controlling the PCH (Section~\ref{sec:dissect}).}
    \label{fig:main_fw}
\vspace{-5mm}
\end{figure*}

\section{Methodology}

\subsection{Hardware Setup}

Our system utilizes a first-generation da Vinci with the dVRK~\cite{dvrk} and a Si model endoscope for enhanced image quality and reduced noise. To control the monopolar current from the electrosurgical unit (ESU), we interfaced the energy cables and console pedals with an Arduino UNO~\cite{oh2024expanded}. Additionally, we calibrated the stereo endoscope to generate 3D point clouds and performed a custom robot arm calibration to minimize dVRK errors~\cite{oh2024framework} (Fig.~\ref{fig:main_fw}(a)). Moreover, we upgraded the implementation platform to ROS2~\cite{ros2022}. 

\subsection{Image/Point Cloud Post-Processing}
We applied the following image-processing steps directly to the tissue segmentation masks provided from the models to design an online version of tracking the boundary of interest and to align the FBF and PCH with the tissues for better performance. First, we used the classical skeletonization algorithm~\cite{lee1994skeleton}, which iteratively sweeps a fixed-size window over the binary mask of the object, removing pixels at each iteration until the image stops changing. Since the gallbladder has an oval-like shape, the 3D skeleton points derived from the point cloud effectively represent its medial surface (yellow points in the figures). The boundary of interest (purple points in the figures) is the interface between the liver and the gallbladder and corresponds to the points on the right side of the skeleton. By applying Principal Component Analysis (PCA)~\cite{abdi2010principal} to the adjacent and skeleton points, we extract the principal axis that points to the right side of the skeleton, aiding in precise boundary alignment.

\begin{figure}[ht]
    \centering
    \includegraphics[trim={0.0 1cm 0.0 0.0}, clip, width=0.45\columnwidth]{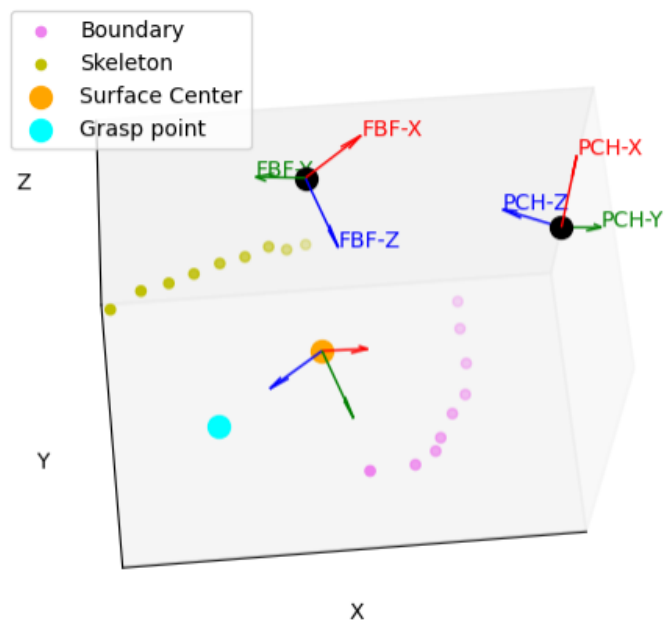}
    \hfill
    \includegraphics[width=0.45\columnwidth]{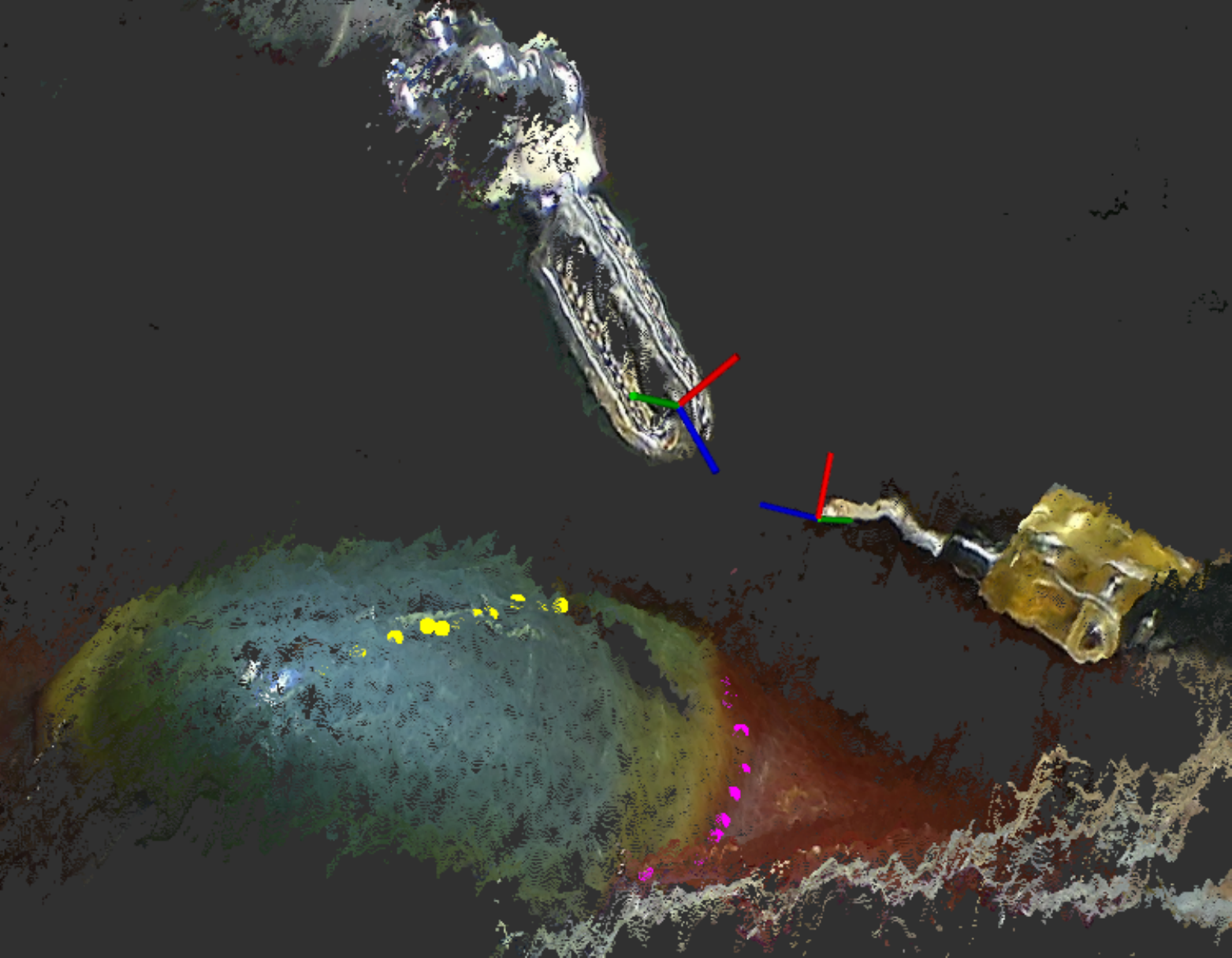}
    \caption{The frames of each instrument and the surface between the skeleton and the boundary of the gallbladder after instrument alignment.}
    \label{fig:align_inst}
\vspace{-2mm}
\end{figure}

\subsection{Grasping}
\label{sec:grasp}

\begin{figure}[ht]
    \centering
    \begin{subfigure}[t]{\columnwidth}
        \centering
        \includegraphics[width=0.5\linewidth]{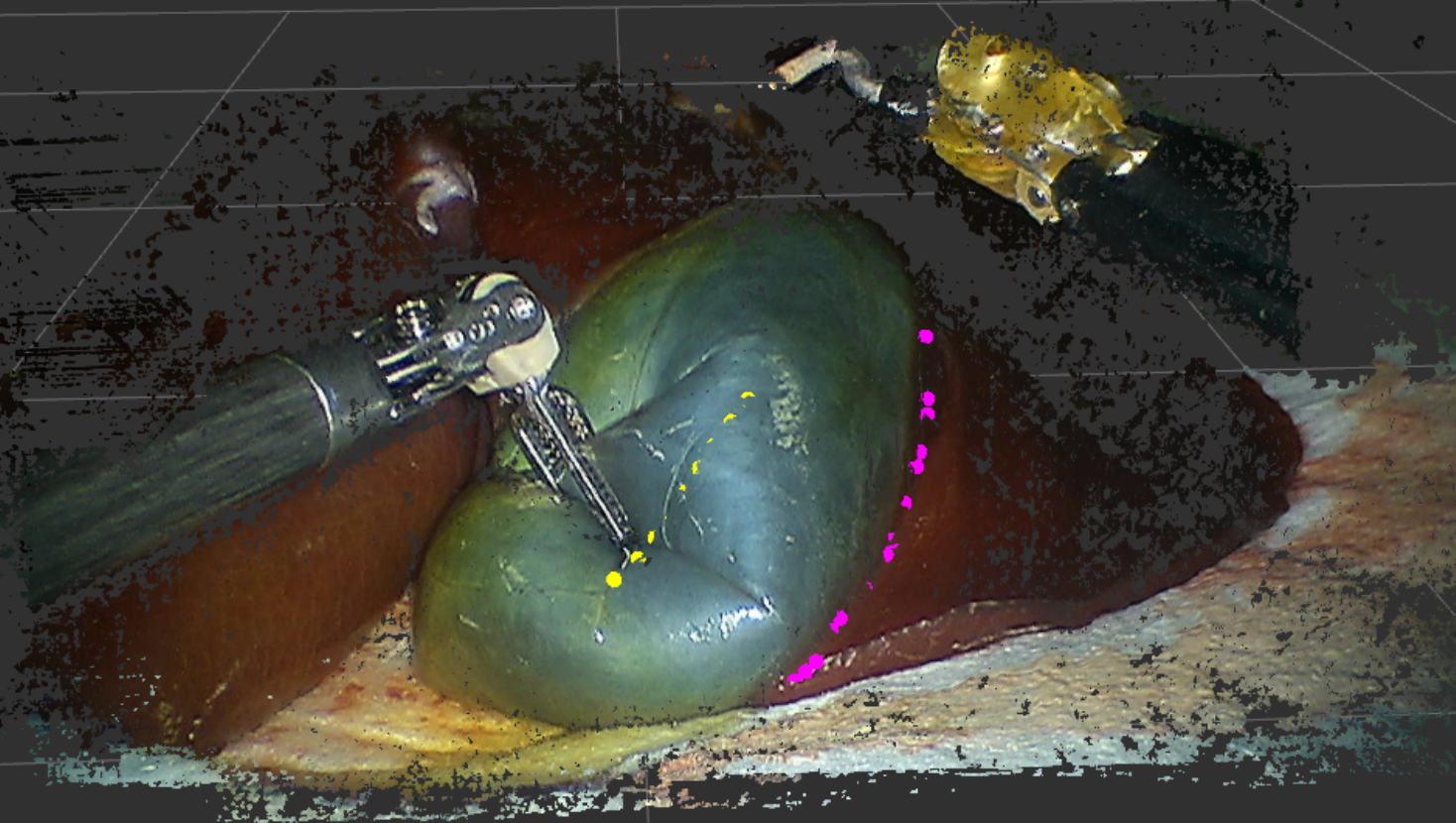}
        \includegraphics[width=0.44\linewidth]{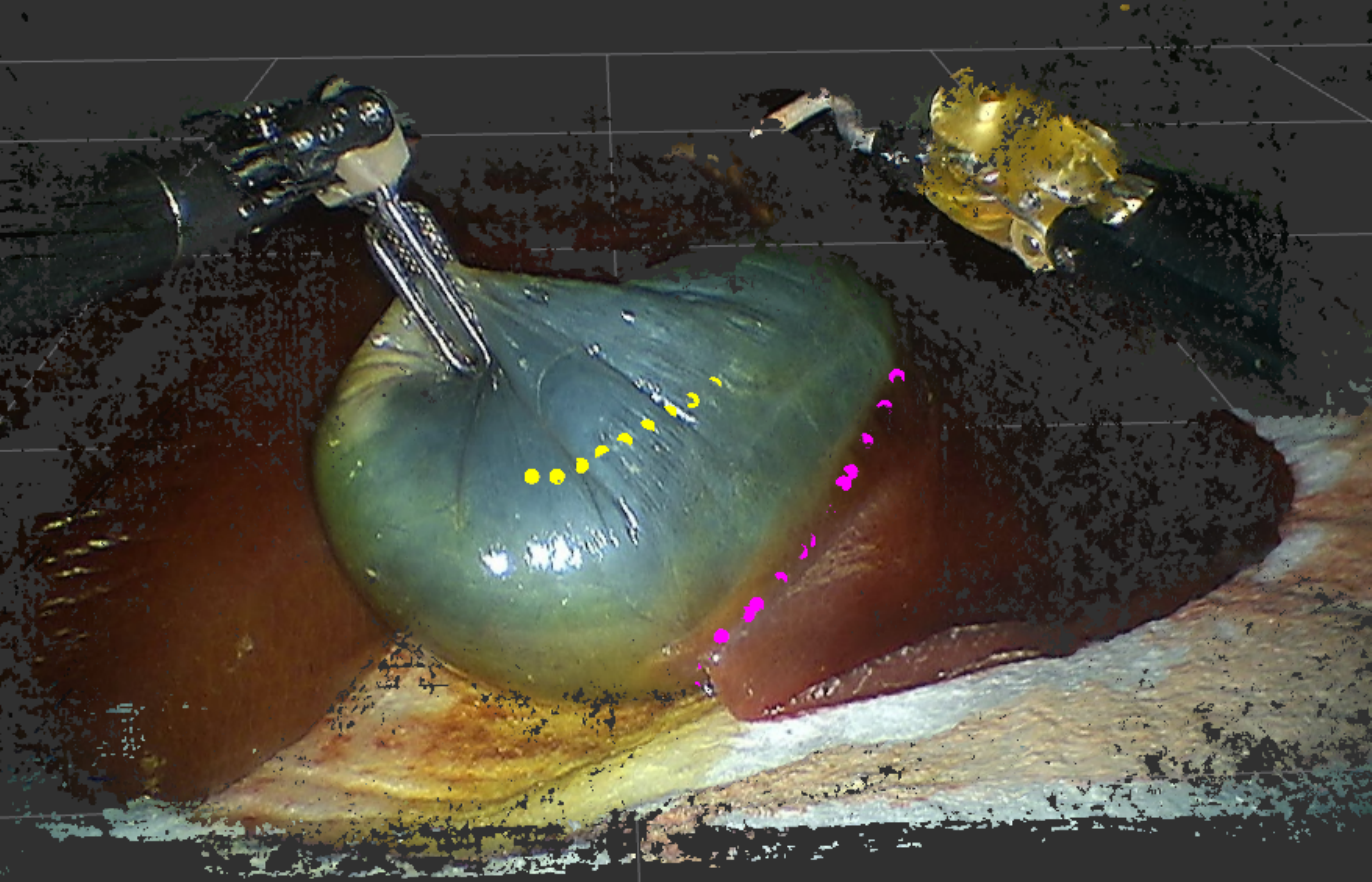}
        \caption{}
        \label{fig:pull_pcl}
    \end{subfigure}
    \begin{subfigure}[t]{\columnwidth}
        \centering
        \includegraphics[width=0.47\linewidth]{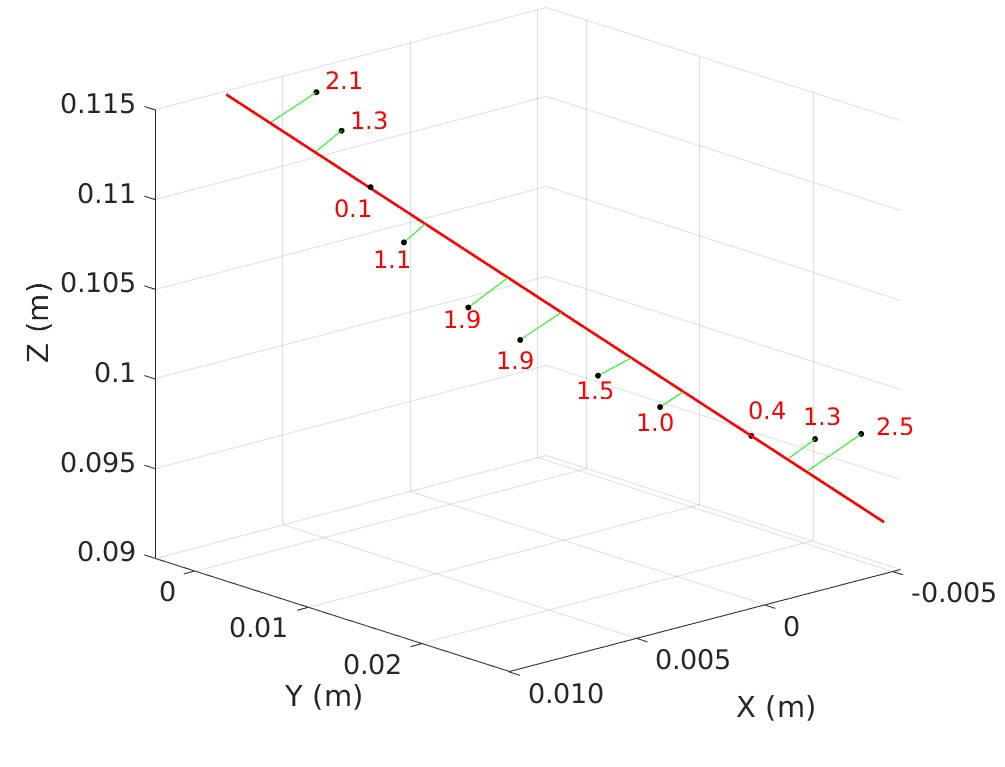}
        \includegraphics[width=0.47\linewidth]{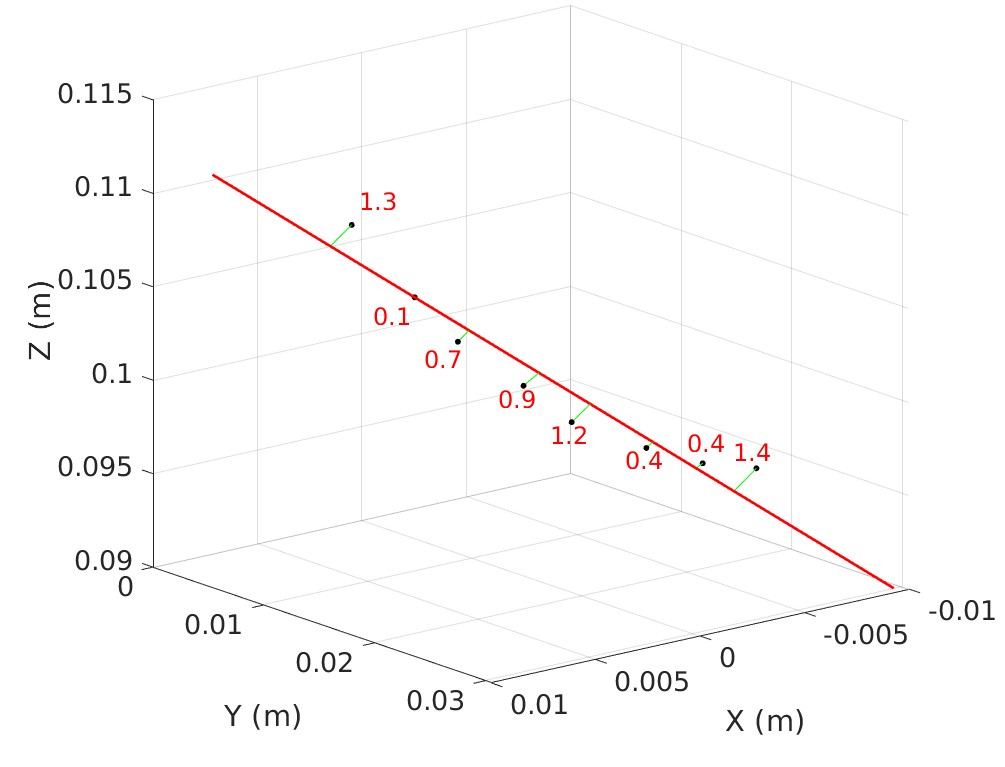}
        \caption{}
        \label{fig:bnd_dev_plot}
    \end{subfigure}
    \hfill
    \begin{subfigure}[t]{\columnwidth}
        \centering
        \includegraphics[width=0.8\linewidth]{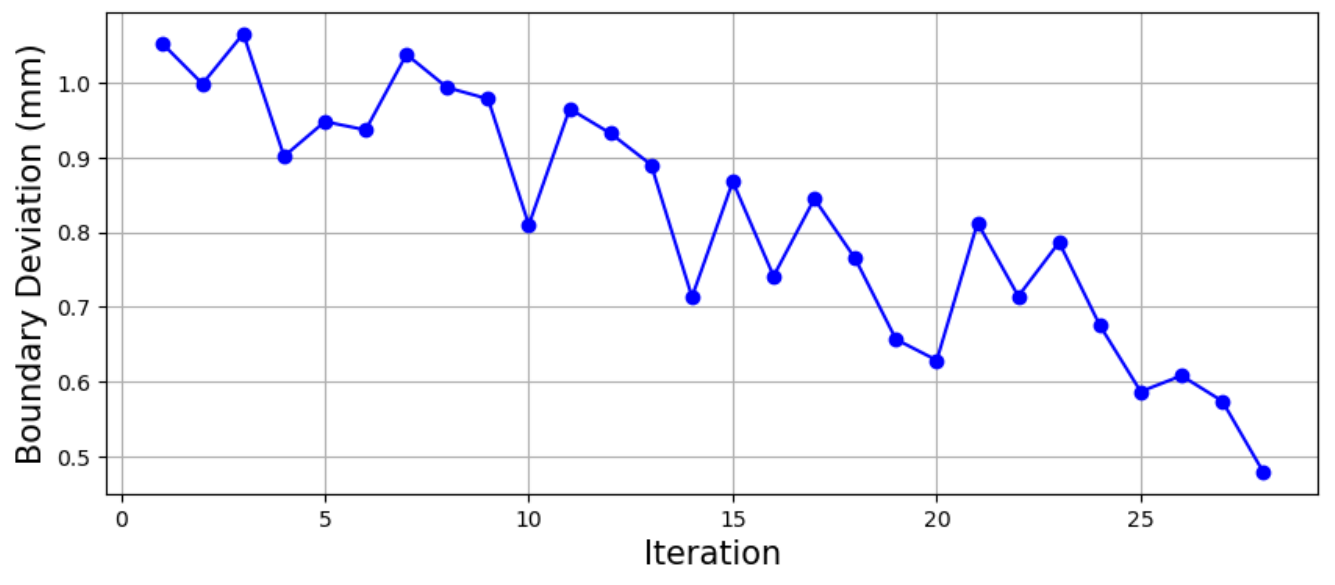}
        \caption{}
        \label{fig:bnd_dev_graph}
    \end{subfigure}
    \caption{(a) Before (left) and after (right) pulling the gallbladder. (b) The deviation of the boundary points to the ideal boundary (red line) before the pull (left) and after the pull (right). (c) Shows the deviation of the boundary points (purple); the lower the deviation, the closer the boundary is to a straight line.}
    \label{fig:pull_gallb}
\vspace{-5mm}
\end{figure}

We divided the grasping task into three steps: align, grasp, and pull. First, we align the pose of the FBF to be parallel to the surface of the gallbladder of interest, which is the region between the boundary and skeleton points obtained from the post-processing step. We apply PCA to the points surrounding this surface to determine the alignment pose. We realigned the principal axes of the gallbladder surface when necessary to ensure that the secondary axis extends from the skeleton to the boundary points and the normal axis of the surface is oriented perpendicular to the gallbladder surface, directed along the viewing direction of the endoscope. This alignment guarantees consistency in orientation across different viewpoints. The resulting three principal axes approximate a local frame attached to the right-side surface of the gallbladder. The FBF frame is oriented based on this local frame (Fig.~\ref{fig:align_inst}). Once aligned, the FBF moves along its negative x-axis, facing through the gallbladder surface, to reach the grasping point between the center of the skeleton and the boundary, offset by a certain distance along the grasping direction. After grasping, the FBF pulls the gallbladder along its negative z-axis, tangential to the right-side surface of the gallbladder, until the boundary becomes nearly linear, where the deviation of the boundary is below $0.5 mm$ (Fig.~\ref{fig:pull_gallb}). This grasping mechanism stabilizes the gallbladder throughout the task, reducing the impact of anatomical variability and sudden tissue deformation.

\subsection{Dissection}
\label{sec:dissect}

Once the grasping and pulling are accomplished, the FBF remains static. Similar to grasping, we first align the PCH to the gallbladder's surface before performing the dissection. The alignment method is identical to grasping, except that the frame axes are mapped based on the PCH's pose rather than the FBF's (Fig.~\ref{fig:align_inst}). Unlike our previous approach, where the PCH followed a fixed, preplanned trajectory, the dissection is now guided online by visual feedback. A target point is dynamically selected from the current tissue boundary, which is updated in real time to reflect changes in the gallbladder’s shape. Initially, the target point is the first in the boundary sequence, ordered clockwise from the center. As the PCH reaches each target, a new one is selected from the updated boundary, choosing the point furthest away (up to 1 cm) in the direction of motion, while maintaining proper alignment. The dissection round ends when no boundary point sufficiently distant (greater than 1 mm) from the current position can be found, ensuring the system adapts continuously to tissue deformation throughout the procedure.
\section{Experimental Results}

\subsection{Instrument Keypoint Detection}

Detectron2~\cite{wu2019detectron2} offers two independent model types: instance object segmentation and keypoint detection. To avoid confusion, we refer to these models as DT2-seg and DT2-kpt. For keypoint prediction with YOLO11, we selected YOLO11l-pose, which delivers performance comparable to its largest variant (YOLO11x-pose) while offering twice the inference speed~\cite{yolo11_ultralytics}. 

\begin{table}[ht]
\centering
\begin{tabular}{l|c|c|c}
\hline\hline
\textbf{Model} & \textbf{Categories}        &\textbf{AP (Bbox.)}& \textbf{AP (Kpt.)} \\ \hline
\multirow{2}{*}{DT2-kpt}    & FBF           &\textbf{77.1} & 94.6          \\
                            & PCH           &74.2 & 98.4          \\
\hline
\multirow{2}{*}{YOLO11l-pose}  & FBF         &66.8          & \textbf{97.2}\\
                               & PCH         &\textbf{85.4} & \textbf{98.6}\\
\hline
\hline
\end{tabular}
\caption{The Average Precision (AP) scores for instrument keypoints detection. }
\label{tab:kpt_comparison}
\vspace{-5mm}
\end{table}

\begin{figure}[ht]
    \centering
    \includegraphics[trim={0.0 3cm 0.0 3cm}, clip, width=1\linewidth]{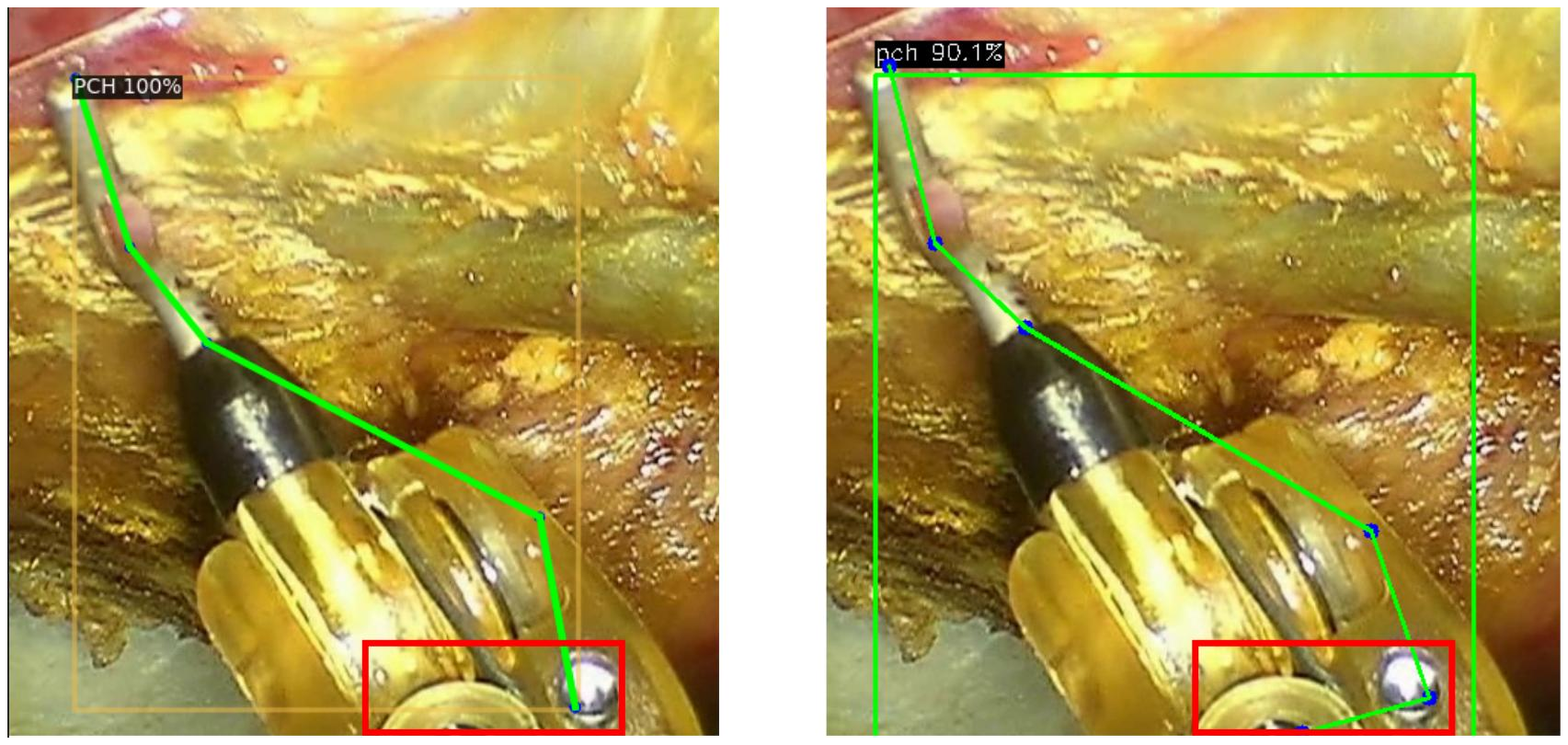}
    \caption{Keypoints prediction for DT2-kpt (left) and YOLO11l-pose (right), this example shows some limitations of the DT2-kpt to detect some keypoints when the tool is on the edges of the image (red rectangle).}
    \label{fig:kpt_comparison}
\vspace{-2mm}
\end{figure}

Table~\ref{tab:kpt_comparison} presents the results of the fine-tuned keypoint detection models. The Average Precision (AP) scores used in this paper are the mAP50-95\footnote{the average of the mean average precision calculated at varying intersection over union (IoU) thresholds, ranging from 0.50 to 0.95}~\cite{mscoco}, which provide a comprehensive view of the model's performance across different levels of detection difficulty. The scores demonstrate that YOLO11l-pose outperforms DT2-kpt. Additionally, Fig.~\ref{fig:kpt_comparison} provides a qualitative comparison, highlighting YOLO11l-pose's improved keypoint localization, particularly in cases where keypoints are partially obscured or near the image frame's edges. These enhancements are crucial in practical applications where occlusion occurs and make keypoint detection accuracy challenging.

\subsection{Tissue Instance Segmentation} 


\begin{table}[ht]
\centering
\begin{tabular}{l|c|c|c}
\hline\hline
\textbf{Model} & \textbf{Categories}       & \textbf{AP (Bbox.)}& \textbf{AP (Seg.)}  \\ \hline
\multirow{3}{*}{DT2-seg}    & Liver       & 61.2 & 83.9                       \\
                            & Gallbladder & 96.2 & 89.8                          \\ 
                            & Liver Bed   & 91.1 & 75.3                          \\
\hline
\multirow{3}{*}{MaskDINO}   & Liver       &55.3& 80.0                          \\
                            & Gallbladder &96.7& \textbf{99.0}                          \\
                            & Liver Bed    &  88.9 & 91.3                      \\
\hline

\multirow{3}{*}{YOLO11l-seg}  & Liver       & \textbf{99.3} & \textbf{98.6}\\
                              & Gallbladder & \textbf{99.5} & 97.7\\
                              & Liver Bed   & \textbf{99.2} & \textbf{93.0}\\
\hline\hline

\end{tabular}
\caption{The Average Precision (AP) scores for each category (Bbox. stands for Bounding box and Seg. for Segmentation).}
\label{tab:seg_comparison}
\vspace{-2mm}
\end{table}

Table~\ref{tab:seg_comparison} presents the AP scores for each model fine-tuned on our new segmentation dataset. The results indicate that segmentation AP scores are more consistent across models, suggesting that the new dataset is more comprehensive than our previous one~\cite{oh2024expanded}. While MaskDINO achieved the highest precision in detecting the gallbladder, YOLO11l-seg outperformed both MaskDINO and DT2-seg in overall performance.

\begin{figure}[ht]
    \centering
    \begin{subfigure}[t]{0.45\columnwidth}
        \centering
        \includegraphics[trim={0.0 0.0 0.0 3.0cm}, clip, width=\linewidth]{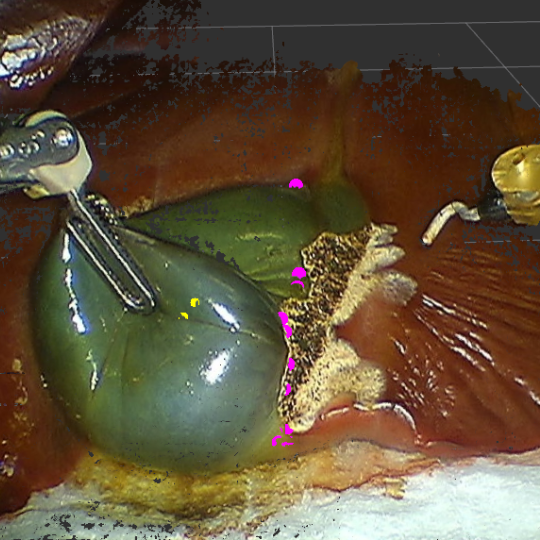}
        \caption{DT2-seg (Old)~\cite{oh2024framework}}
        \label{fig:bnd_after_dissect_old}
    \end{subfigure}
    \hfill
    \begin{subfigure}[t]{0.45\columnwidth}
        \centering
        \includegraphics[trim={0.0 0.0 0.0 3.0cm}, clip, width=\linewidth]{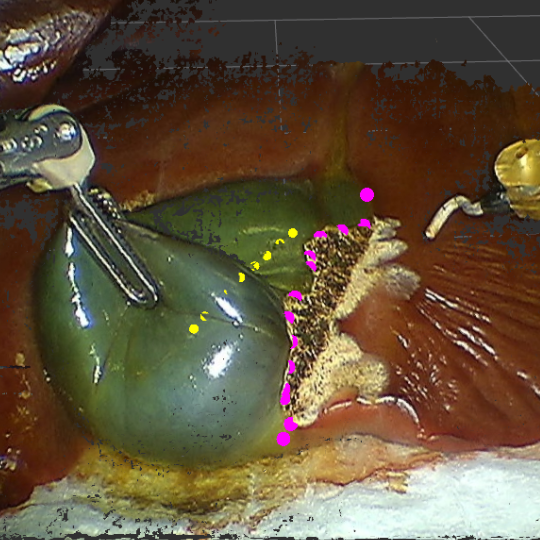}
        \caption{YOLO11}
        \label{fig:bnd_after_dissect_new}
    \end{subfigure}
    \caption{Comparing the performance between (a) the model we used in our previous work and (b) the recent YOLO11 on detecting the bounding after a few rounds of dissection.}
    \label{fig:bnd_after_dissect}
\vspace{-5mm}
\end{figure}

Both DT2-seg and MaskDINO struggled with detecting the \textit{liver}, likely due to its similarity to the \textit{liver bed}, which is part of the liver's surface layer. In contrast, YOLO11l-seg demonstrated the ability to differentiate between these two tissues, as reflected in its AP scores. The limitations of DT2-seg and MaskDINO may not be critical during the initial stages of dissection when the liver bed is not yet exposed. However, their reduced accuracy could impact performance in later stages, particularly after the liver has cauterization marks. 

Nevertheless, Fig.~\ref{fig:bnd_after_dissect} highlights the necessity of the new class and its impact on boundary determination after multiple rounds of dissection to advance towards actual robotic cholecystectomy. In Fig~\ref{fig:bnd_after_dissect_old}, the previous model's estimations show a disconnected boundary and difficulties detecting the full gallbladder, displaying an incomplete skeleton. In contrast, Fig.~\ref{fig:bnd_after_dissect_new} demonstrates the improved model's ability to accurately detect the boundary and the entire gallbladder with a complete skeleton.

\subsection{Procedure Performance}

Fig.~\ref{fig:align_inst} and Fig.~\ref{fig:pull_gallb} illustrate the stages of gallbladder grasping. The success of the grasping phase does not heavily depend on the models used; instead, the primary objective is to ensure that the boundary is sufficiently stretched for dissection and that the FBF remains stable throughout the procedure. This evaluation also serves as an ablation analysis: our previous system~\cite{oh2024framework} lacked grasping, alignment, and used older perception models, enabling us to isolate the impact of each updated module in the current framework. Consequently, our focus is primarily on evaluating the dissection performance after the gallbladder has been adequately stretched, allowing for a direct comparison with results from our previous work. The FBF was not repositioned after the initial grasping to ensure consistency across trials. For Detectron2, we used DT2-seg and DT2-kpt to maintain the previous experimental setup, with both models fine-tuned on the new dataset. We then compared their performance with both our prior results and state-of-the-art models.

\begin{figure}[ht]
    \centering
    \begin{subfigure}[t]{0.48\columnwidth}
        \centering
        \includegraphics[trim={0.0 0.0 0.0 0.0cm}, clip, width=\linewidth]{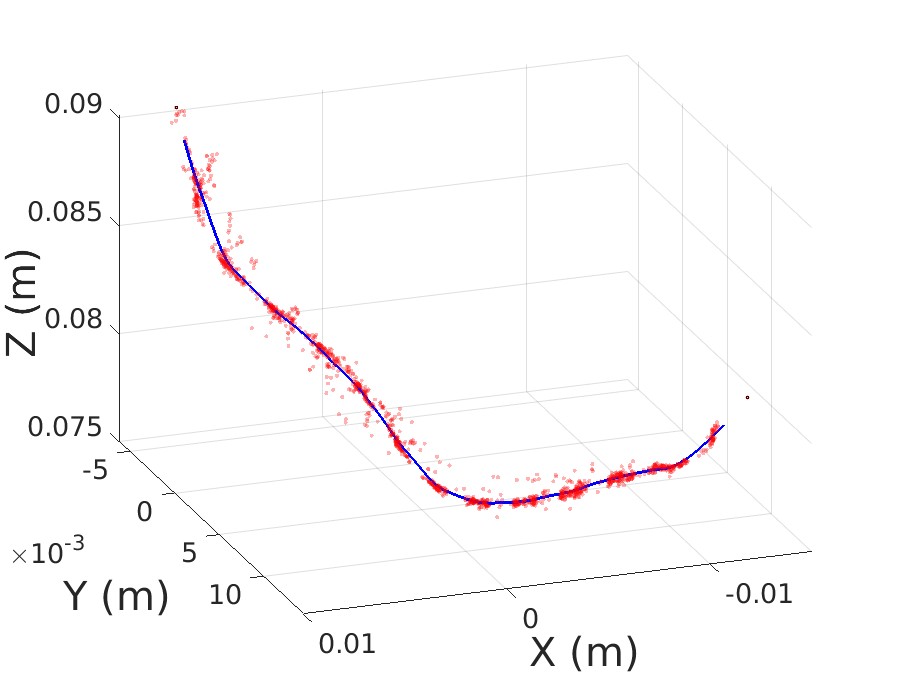}
        \caption{YOLO11l-seg}
        \label{fig:bnd_distribution_yolo}
    \end{subfigure}
    \begin{subfigure}[t]{0.48\columnwidth}
        \centering
        \includegraphics[trim={0.0 0.0 0.0 0.0cm}, clip, width=\linewidth]{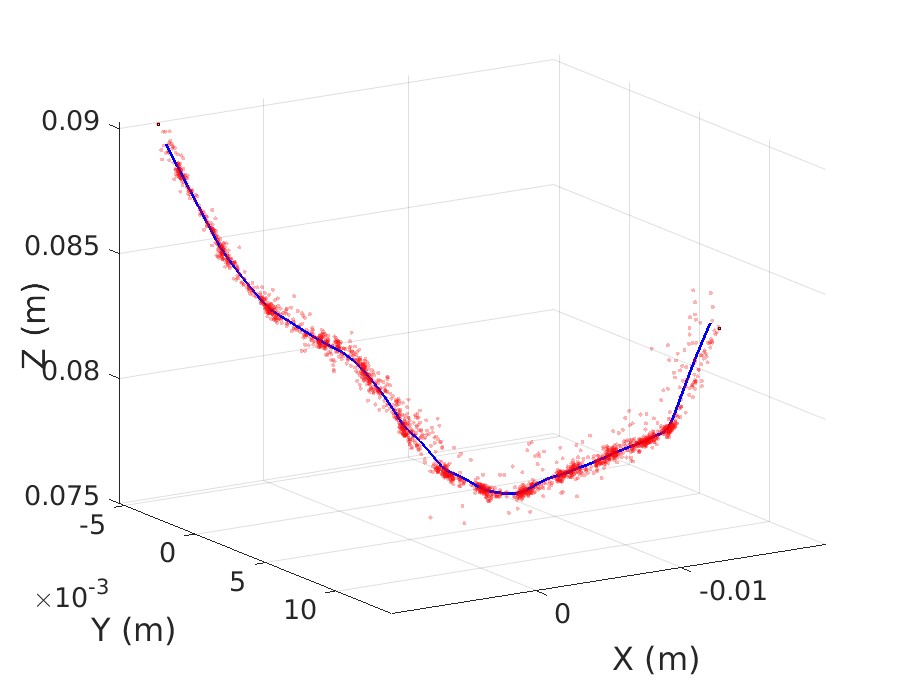}
        \caption{DT2-seg}
        \label{fig:bnd_distribution_dt2}
    \end{subfigure}
    \caption{Samples of boundary points in a single trial, where (a) is from YOLO11l-seg (Trial 1 in Tab. \ref{tab:procedure_results}) and (b) is from DT2-seg (Trial 2).}
    \label{fig:bnd_distribution}
\vspace{-3mm}
\end{figure}

Fig.~\ref{fig:bnd_distribution} presents the recorded boundary points from one of the trials, comparing YOLO11l to DT2,  demonstrating that YOLO's boundary remained stable over time. The results indicate that the dissection process did not significantly alter the boundary, even though boundary updates were performed in real-time. This validates the effectiveness of our grasping strategy in maintaining boundary stability during dissection, which was a key limitation in our previous study. We fitted a cubic spline (blue lines in Fig.~\ref{fig:bnd_distribution}) to the recorded boundary points to quantify boundary consistency, computing the residual mean squared error (RMSE) between the points and the fitted curve, as shown in Table~\ref{tab:procedure_results}. Since the boundary points were fixed in our previous work, we excluded those results from the table. Additionally, we measured the total travel distance of the PCH during dissection and the time required to complete one full boundary dissection, including data from our prior study. We do not explicitly report a task success rate or boundary coverage metric, as the dissection boundary is adaptively updated in real time to account for tissue deformation while performing the task.

\begin{table}[ht]
    \centering
    \resizebox{\columnwidth}{!}{%
    \begin{tabular}{l|c|c|c|c}
       \hline\hline
       \textbf{Model (Seg. + Kpt.)} & \textbf{Trial} & \textbf{RMSE (mm)} & \textbf{Distance (mm)} & \textbf{Duration (s)} \\
       \hline
       \multirow{6}{*}{\shortstack{DT2-seg \\ + \\ DT2-kpt \\ (Old)}}
            & 1 & - & 60.3 & 104  \\
            & 2 & - & 51.3 & 117  \\
            & 3 & - & 28.7 & 79  \\
            & 4 & - & 36.3 & 92  \\
            & 5 & - & 38.3 & 97  \\
            & 6 & - & 45.5 & 116  \\
        \hline
        Mean & & - & 43.4 & 100.8   \\
        Std. Dev. &&-&11.3&14.6\\
        \hline\hline
        \multirow{5}{*}{\shortstack{DT2-seg \\ + \\ DT2-kpt \\ (New)}}
            & 1 & 0.42 & 50.5 & 29  \\
            & 2 & 0.78 & 51.1 & 28  \\
            & 3 & 0.50 & 65.2 & 35 \\
            & 4 & 0.43 & 54.3 & 30  \\
            & 5 & 0.46 & 82.9 & 40  \\
            
        \hline
        Mean & & 0.51 & 60.8 & 32.4   \\
        Std. Dev. && 0.13&13.7&5.3\\   
        \hline\hline
        \multirow{5}{*}{\shortstack{MaskDINO \\ + \\ YOLO11l-pose}}
            & 1 & 0.51 & 56.1 & 60   \\
            & 2 & 0.63 & 46.3 & 30   \\
            & 3 & 0.53 & 65.5 & 59   \\
            & 4 & 0.49 & 69.3 & 64   \\
            & 5 & 0.49 & 62.3 & 65  \\
        \hline
        Mean & & 0.53 & 59.9 & 55.4  \\
        Std. Dev. && 0.05&9.0&14.5\\ 
        \hline\hline
        \multirow{5}{*}{\shortstack{YOLO11l-seg \\ + \\ YOLO11l-pose}}
            & 1 & 0.47 & 54.9 & 31  \\
            & 2 & 0.52 & 47.2 & 29  \\
            & 3 & 0.47 & 56.0 & 32  \\
            & 4 & 0.50 & 54.9 & 31  \\
            & 5 & 0.51 & 48.0 & 29  \\
         \hline
         Mean & & \textbf{0.49} & 52.2 & \textbf{30.6} \\
         Std. Dev. &&\textbf{0.02}&\textbf{4.2}&\textbf{1.3}\\
         \hline\hline
    \end{tabular}%
    }
    \caption{Performance metrics across different models (Seg. + Kpt.) and trials. Residual mean squared error (RMSE) of the boundaries (Fig.~\ref{fig:bnd_distribution}) and the average and standard deviation values across trials are provided in the last row for each model.}
    \label{tab:procedure_results}
\end{table}

As shown in Table~\ref{tab:procedure_results}, the YOLO11 model achieved the lowest RMSE, maintaining consistency across trials. In contrast, other models exhibited occasional RMSE spikes even though we removed the outliers for DT2-seg and MaskDINO before the analysis. A similar trend was observed for the travel distance of the PCH, where YOLO11 demonstrated stable performance while other models experienced variability, shown by a higher variance in all measurements for DT2 and MaskDINO. These discrepancies can be attributed to the limited predictive accuracy of DT2-kpt (as illustrated in Fig.~\ref{fig:kpt_comparison}) and noise in the point clouds surrounding the instrument, which led to wrong instrument tip positions. This, in turn, caused oscillatory movements of the robotic arm, increasing both travel distance and the duration of the procedure.

Furthermore, the dissection speed significantly improved compared to our previous DT2 (Old) and current DT2 (New) results. This improvement can be attributed to two main factors. First, the updated DT2-kpt model, trained on a more recent dataset, provided more accurate instrument keypoint predictions than the outdated version used in previous work. Second, the alignment of the PCH being orthogonal to the gallbladder surface before dissection minimized contact between the PCH and the liver, preventing the instrument from getting stuck at specific locations.

\section{Conclusion}

In this work, we presented a comprehensive framework for automated robotic dissection in cholecystectomy. Building on our previous efforts, we introduced a novel segmentation dataset derived from a more realistic robotic cholecystectomy dataset (CRCD). Additionally, incorporating the liver bed class improved boundary tracking even after cauterization. We fine-tuned state-of-the-art instance segmentation models, including MaskDINO and YOLO11, and compared their performance with our previously used model, Detectron2. 

Furthermore, we transitioned from offline to online boundary point extraction, improving the methodology to ensure robustness under challenging surgical conditions, including multiple dissections, tissue color changes over time, and gallbladder size and shape variations. Additionally, we aligned the surgical instruments to the gallbladder surface before each action, enhancing task performance. We also introduced an automated bimanual manipulation strategy that integrates grasping and pulling actions, which are crucial for maintaining optimal tissue tension and compensating for shifts caused by energy application. To validate our framework, we conducted ex vivo evaluations on porcine livers, demonstrating improved surgical instrument localization and precise dissection, even after tissue deformation.

While this extended framework improved upon our previous work and addressed key limitations, further refinements are necessary. Through our work with the CRCD, we observed that surgeons employ different dissection techniques. One approach, as implemented in our current framework, involves following the boundary after stretching the boundary between the gallbladder and liver. However, another common technique involves using the tip of the PCH to hook the boundary and either pulling until the tissues detach or applying energy if the tissue remains too thick after pulling. This method minimizes damage to the liver during energy application, reducing the risk of severe postoperative complications~\cite{JAUNOO201015}.


Although the keypoint detection has improved from our previous work, it still suffers in certain poses or situations during the task. To address this, we plan to further augment the dataset for keypoints using the CRCD dataset, similar to how we expanded the segmentation dataset. In parallel, we are developing a robustness strategy in which the endoscope actively follows the instrument tip, ensuring it remains in the center of the image to mitigate occlusion and boundary visibility issues during keypoint detection. We also aim to refine our control strategy to fully separate the gallbladder from the liver, accomplishing multiple rounds of pulling and dissecting the gallbladder autonomously and comparing the performance with the surgeons. 

Overall, our contributions mark a significant advancement in robotic-assisted surgery, providing a robust foundation for the future development of autonomous dissection techniques in minimally invasive procedures.

\nocite{*}
\bibliographystyle{IEEEtran}
\bibliography{iros_2025}

\end{document}